%% file: main.tex
\documentclass[sigconf]{acmart}
\input{meta/packages.tex}
\input{meta/copyright.tex}

\input{meta/commands.tex}
\usepackage{subcaption}
\usepackage{booktabs}
\usepackage{makecell}   
\usepackage{adjustbox}

\renewcommand\footnotetextcopyrightpermission[1]{} 
\begin{document}

\title{Training Machine Learning Models on Human Spatio-temporal Mobility Data: An Experimental Study [Experiment Paper]}
\renewcommand{\shorttitle}{Training Machine Learning Models on Human Spatio-temporal Mobility Data}
\input{meta/authors.tex}
\renewcommand{\shortauthors}{Liu et al.}
\begin{abstract}
    \input{content/abstract.tex}
\end{abstract}

\input{meta/ccs.tex}
\vspace{-0.6cm}
\vspace{-0.6cm}

\maketitle

\section{Introduction}
\label{sec:introduction}
\input{content/introduction.tex}

\vspace{-0.2cm}
\section{Background and Related Work}
\vspace{-0.0cm}
\label{sec:related_works}
\input{content/related_works.tex}

\vspace{-0.2cm}
\section{Machine learning on Human Mobility Prediction}
\label{sec:preliminary}
\input{content/preliminary.tex}

\vspace{-0.65cm}
\section{Model and Experimental Setting}\vspace{-0.1cm}
\label{sec:methodology}
\input{content/methodology.tex}

\section{Experimental Results and Analysis}
\label{sec:results}
\input{content/results.tex}

\vspace{-0.2cm}
\section{Conclusions and Future Work}
\label{sec:conclusion}
\input{content/conclusion.tex}

\bibliographystyle{ACM-Reference-Format}
\bibliography{refs/main}

\end{document}

%% file: meta/packages.tex
\usepackage{graphicx}
\usepackage{hyperref}
\usepackage{listings}
\usepackage{booktabs}
\usepackage{multirow}
\usepackage{amsmath}
\usepackage[most]{tcolorbox}
\usepackage{xcolor}
\usepackage{booktabs} 
\usepackage{amsmath}
\usepackage{amsthm}
\usepackage{multirow,tabularx,ragged2e}
\usepackage{adjustbox}
\usepackage{graphicx}
\usepackage{hyperref}
\usepackage{enumitem}
\usepackage{lipsum}
\usepackage{multirow}

%% file: meta/copyright.tex
\copyrightyear{2025}
\acmYear{2025}
\setcopyright{acmcopyright}
\acmConference[SIGSPATIAL'25]{33rd ACM SIGSPATIAL International Conference on Advances in Geographic Information Systems}{November 03--November 06, 2025}{Minneapolis, MN, USA}
\acmISBN{978-1-4503-XXXX-X/18/06}

%% file: meta/commands.tex
\def\andreas#1{{\footnotesize{\sc\textcolor{red}{}}}}
\def\andi#1{{\footnotesize{\sc\textcolor{red}{}}}}
 \def\hossein#1{{\footnotesize{\sc\textcolor{blue}{}}}}
 \def\lance#1{{\footnotesize{\sc\textcolor{blue}{}}}}
\def\lance#1{{\footnotesize{\sc\textcolor{blue}{}}}}
\def\yl#1{{\footnotesize{\sc\textcolor{violet}{}}}}
\def\rk#1{{\footnotesize{\sc\textcolor{orange}{}}}}
\def\joon#1{{\footnotesize{\sc\textcolor{orange}{}}}}

%% file: meta/authors.tex
\author{Yueyang Liu}
\orcid{0000-0001-5894-8740}
\affiliation{%
\institution{Emory University, Atlanta, USA}
  \city{}
  \state{}
  \country{}
}
\email{yueyang.liu@emory.edu}

\author{Lance Kennedy}
\orcid{0009-0004-6815-2219}
\affiliation{%
  \institution{Emory University, Atlanta, USA}
  \city{}
  \state{}
  \country{}
}
\email{lance.kennedy@emory.edu}

\author{Ruochen Kong}
\orcid{0009-0006-0329-8019}
\affiliation{%
  \institution{Emory University, Atlanta, USA}
  \city{}
  \state{}
  \country{}
}
\email{ruochen.kong@emory.edu}

\author{Joon-Seok Kim}
\orcid{0000-0001-9963-6698}
\affiliation{%
  \institution{Emory University, Atlanta, USA}
  \city{}
  \state{}
  \country{}
}
\email{joonseok.kim@emory.edu}

\author{Andreas Z{\"u}fle}
\orcid{0000-0001-7001-4123}
\affiliation{%
  \institution{Emory University, Atlanta, USA}
  \city{}
  \state{}
  \country{}
}
\email{azufle@emory.edu}

\renewcommand{\shortauthors}{Liu et al.}

%% file: content/abstract.tex
Individual-level human mobility prediction has emerged as a significant topic of research with applications in infectious disease monitoring, child, and elderly care. Existing studies predominantly focus on the microscopic aspects of human trajectories: such as predicting short-term trajectories or the next location visited, while offering limited attention to macro-level mobility patterns and the corresponding life routines. 
In this paper, we focus on an underexplored problem in human mobility prediction: determining the best practices to train a machine learning model using historical data to forecast an individuals complete trajectory over the next days and weeks.
In this experiment paper, we undertake a comprehensive experimental analysis of diverse models, parameter configurations, and training strategies, accompanied by an in-depth examination of the statistical distribution inherent in human mobility patterns.
Our empirical evaluations encompass both Long Short-Term Memory and Transformer-based architectures, and further investigate how incorporating individual life patterns can enhance the effectiveness of the prediction. We show that explicitly including semantic information such as day-of-the-week and user-specific historical information can help the model better understand individual patterns of life and improve predictions. 
Moreover, since the absence of explicit user information is often missing due to user privacy, we show that the sampling of users may exacerbate data skewness and result in a substantial loss in predictive accuracy. To mitigate data imbalance and preserve diversity, we apply user semantic clustering with stratified sampling to ensure that the sampled dataset remains representative. Our results further show that small-batch stochastic gradient optimization improves model performance, especially when human mobility training data is limited.

%
%

%% file: meta/ccs.tex
\ccsdesc[500]{Information systems~Geographic information systems}
\ccsdesc[500]{Information systems~Location based services}

%% file: content/introduction.tex
Individual-level human mobility prediction has predominantly been constrained to the microscopic scale, focusing on short-term behavioral patterns, such as the next visited location, employing methods such as Markov models~\cite{ye2013s,cheng2013you,he2016inferring}, collaborative filtering~\cite{lian2013collaborative}, Wavelet Transform~\cite{assam2014check}, Attention~\cite{luo2021stan}, and Recurrent Neural Networks~\cite{liu2016predicting}.
%
But predicting the mobility of individuals over days and weeks remains a largely underexplored problem. One of the primary obstacles towards long-term individual-level mobility prediction is the scarcity of large-scale, high-fidelity human mobility datasets due to privacy concerns and the cost of data collection \cite{8356232}. For example, the widely used GeoLife dataset~\cite{zheng2011geolife} includes trajectory data for only 182 individual users.
%
%
A further limitation of this dataset is the fragmented nature of the trajectories—very few users possess long-term, continuous records suitable for extended trajectory prediction tasks. Similarly, while the Gowalla dataset~\cite{10.1145/2020408.2020579} contains a total of 196,591 users and approximately 6,442,890 check-ins, the average number of check-ins per user is only around 32, which is insufficient for capturing long-term behavioral patterns. Other datasets, such as the T-Drive Taxi dataset~\cite{zheng2011t-drive}, although rich in GPS traces, capture taxi cabs, not individual humans, and thus do not capture any individual's patterns of life. To the best of our knowledge, the largest (and publicly available) dataset of individual human locations is Foursquare data~\cite{6844862}, which captures ``check-ins'' of users at places that are part of the Foursquare Location-Based Social Network application. For this dataset, a prior study in~\cite{10.1007/s10707-016-0279-5} has shown that these check-ins are highly unpredictable and show that at least 27\% of check-ins are theoretically impossible to predict. 

Despite these data limitations, predicting individual-level human mobility has many potential applications: It allows us to understand where individuals should (not) be at different times. For instance, sudden deviations of many individuals' normal patterns of life may indicate anomalies, such as the outbreak of an infectious disease long before these individuals may report symptoms and get tested. Deviation of an elderly individual's mobility may indicate being lost or confused and a child's deviation from normal may point towards skipping school or a possible abduction.


Yet, critical challenges such as how to exploit the structure and semantics of mobility data and how to train long-term trajectories prediction machine learning models remain largely unresolved.
This paper aims to address this gap by exploring machine learning training methodologies for accurately predicting a user’s point-of-interest (PoI) check-in sequences over a multi-day horizon, based on limited historical spatial-temporal data.

In this experiment paper, we utilize each user’s recent mobility history—for example, trajectory data from the preceding month—to forecast future behavior. 
Specifically, we focus on predicting the full monthly trajectory by initially modeling users life pattern and PoI semantics. 
To utilize the existing real-world dataset, the task is formulated as the prediction of the likelihood that a user will be present at a particular location during specific time intervals.
To address this problem, we explore several candidate models that are potentially well-suited for trajectory sequence prediction, including both recurrent and attention-based architectures. 
Furthermore, for parts of the dataset that include contextual information about PoI surroundings, we conduct a comparative study to evaluate the contribution of different data modalities to overall model performance.


We observe that the assumption of independent and identically distributed (IID) data, commonly made in many machine learning models~\cite{dundar2007learning,10192897}, is not satisfied in trajectory data as different groups of people exhibit vastly different distributions of places they visit. For example, children are more likely to visit schools and elderly are more likely to visit healthcare facilities. In addition, trajectory data is inherently biased, as certain demographic populations may be over- or underrepresented in the set of users that are captured in datasets like FourSquare and GoWalla. These biases may arise from sensor heterogeneity, user behavior, or contextual factors. At a more practical level, external conditions such as traffic congestion, severe weather, or public holidays can induce correlated mobility behaviors among otherwise unrelated users—for example, co-traveling patterns during mass events. These problems of data bias and non-independent and identical distributions in trajectory data may lead to overfitting of models. For example, a dataset having mostly young users may fail at predicting a small number of old users' mobility. To address this limitation, we propose to cluster individual users semantically, and perform a stratified sampling to ensure that machine learning models see sufficient examples of different semantic classes of users.



This experiments paper makes the following contributions: 
\vspace{-0.1cm}
\begin{enumerate}[leftmargin=*]
\item  A Unified Spatio-Temporal Prediction Framework: We proposed a comprehensive framework that integrates temporal segmentation, user semantic embeddings, and historical check-in frequency through a fusion layer, enabling accurate full-day trajectory prediction. 

\item Stratified User Sampling: We investigate the impact of non-independent and identically distributed (non-IID) structures inherent in human mobility data. We design and evaluate sampling mechanisms, including (1) user-level clustering to introduce the learning models to distributional shifts and (2) stratified sampling, to address the non-IID nature of human mobility data, improving both generalization and training stability. 

\item Investigation of Training Stability Under Varying Batch Sizes: We conduct a comprehensive series of experiments using small-batch stochastic gradient optimization. The results demonstrate that conventional human mobility models are sensitive to batch size, which directly refer to model generalization performance. Furthermore, our findings indicate that GEO-BLEU offers more detailed and informative insights for evaluating model generalization compared to traditional metrics.

\item	Comparative Study on Spatio-Temporal Mobility Datasets:
To better understand the contribution of each model component, we provide a comparative study. This study emphasizes the critical role of modeling individual life patterns and demonstrates that eliminating irrelevant or redundant inputs can degrade predictive performance. Specifically, we evaluate the impact of three key factors: (1) extended temporal information (e.g., rush hour segmentation), (2) user semantic embeddings derived from check-in history and PoI metadata, and (3) historical life pattern representations based on long-term behavioral regularities. 


\end{enumerate}
To reproduce our experiments, all source code and data is available at: https://github.com/alex-cse/Training-Machine-Learning-Models-on-Human-Mobility-Data.

%% file: content/related_works.tex
\begin{table*}[t]
    \centering
    \scriptsize
    \caption{ Template Structure of Human Spatio-Temporal Trajectory Data Derived from the Foursquare Dataset. 
    \vspace{-0.1cm}}
    \begin{adjustbox}{width=1.0\linewidth,center}
\begin{tabular}{rlllrrrl}
\toprule
user\_id & venue\_id & venue\_category\_id & venue\_name & lat & lon & time\_zone\_offset & utc\_time \\
\midrule
1541 & 4f0fd5a8e4b03856eeb6c8cb & 4bf58dd8d48988d10c951735 & Cosmetics Shop & 35.71 & 139.62 & 540 & Tue Apr 03 18:17:18 +0000 2012 \\
868 & 4b7b884ff964a5207d662fe3 & 4bf58dd8d48988d1d1941735 & Ramen /  Noodle House & 35.72 & 139.80 & 540 & Tue Apr 03 18:22:04 +0000 2012 \\
114 & 4c16fdda96040f477cc473a5 & 4d954b0ea243a5684a65b473 & Convenience Store & 35.71 & 139.48 & 540 & Tue Apr 03 19:12:07 +0000 2012 \\
\bottomrule
\end{tabular}
    \end{adjustbox}
    \label{table:datasample}
    \vspace{-0.1cm}
\end{table*}

Here, 
in Section~\ref{sec:related_db} we first introduce several widely used human mobility check-in datasets, providing detailed information on their data sources, temporal coverage, and the challenges associated with acquiring such data—particularly due to privacy constraints, collection costs, and the rarity of long-term, fine-grained mobility records.
Subsequently, in Section~\ref{sec:related_paper}, we review several existing studies on human mobility prediction and in Section~\ref{sec:related_batch},
we discuss some existing findings regarding the use of small batch sizes in machine learning training and how small-batch stochastic gradient optimization strategy contributes to improved model performance.
After that, we explore the non-independent and identically distributed (non-IID) characteristics in Section~\ref{sec:related_iid}, where we examine the implications of non-IID sampling and how those characteristics impact human mobility prediction models.


%
%
\vspace{-0.2cm}
\subsection{Human Mobility Datasets}\label{sec:related_db}
In this section, we introduce several widely used datasets commonly employed in human mobility prediction tasks. Specifically, this paper focuses on long-term, fine-grained human mobility prediction, where the goal is to forecast individual users’ future trajectories over extended time horizons. In this context, vision-based datasets such as UCY and ETH, which are primarily designed for short-term crowd behavior modeling, as well as aggregated group-level datasets like PeMS, TaxiBJ, and Yahoo! Bousai, which capture human flow or traffic patterns, are not directly relevant to our study.
Under this scenario, the availability of high-quality datasets is exceedingly rare due to significant privacy concerns and the high costs associated with data collection.

To advance our research, we examine several open-access datasets that primarily consist of check-in records from social networking platforms, as well as GPS traces collected via smartphones or in-vehicle GPS devices.
\newpage

\subsubsection{The GeoLife dataset}\cite{zheng2011geolife} This dataset contains trajectories collected from 182 users over a period exceeding three years. Most trajectories are recorded with high spatial and temporal resolution, often at the level of seconds or meters. Each trajectory point is annotated with latitude, longitude, altitude, and timestamp, enabling fine-grained analysis of user mobility.

\begin{figure}[t]
    \centering
    \includegraphics[width=1\linewidth,trim=0cm 0cm 0.0cm 0.0cm,clip]{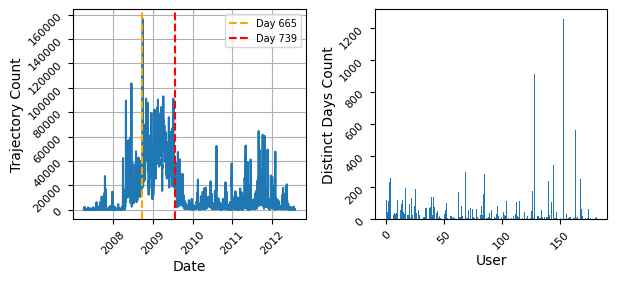}
    \vspace{-0.1cm}
    \captionsetup{width=1.\linewidth}
    \caption{GeoLife - Daily Trajectory Number and User Trajectory Length (count by day).\vspace{-0.2cm}}
    \label{fig:geo_life}
    \vspace{-0.1cm}
\end{figure}

Upon reviewing the dataset, we observe that on more than 50\% of days, the total number of trajectory records is fewer than 3,400. At the user level, 75\% of the users are active on only 63 days throughout the entire dataset. The overall distribution of daily trajectory counts and individual user trajectory lengths is visualized in Figure~\ref{fig:geo_life}. To isolate a high-density subset for long-term prediction, we restricted our analysis to the UTM Zone 50N region and applied a density function to identify a contiguous 300-day interval (from day 535 to 834) that contains the majority of the trajectory data. This subset comprises 11,792,656 records but includes only 72 unique users.  

\subsubsection{Foursquare dataset}\cite{6844862} This dataset comprises check-in records collected over approximately 10 months from two major metropolitan areas: New York City and Tokyo. It includes a total of 227,428 check-ins in New York City and 573,703 check-ins in Tokyo. Each check-in is annotated with a timestamp, GPS coordinates, and a semantic label representing a fine-grained venue category. A sample of the dataset is provided in Table~\ref{table:datasample}.
\begin{figure}[t]
    \centering
    \includegraphics[width=1\linewidth,trim=0cm 0cm 0.0cm 0.0cm,clip]{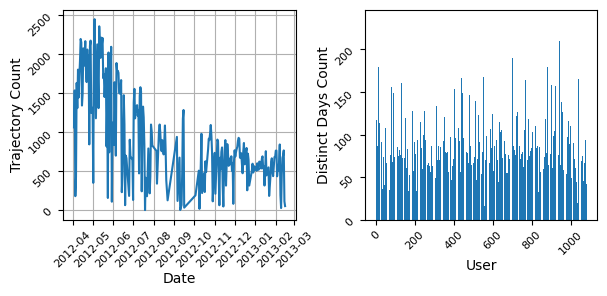}
    \vspace{-0.1cm}
    \captionsetup{width=1.\linewidth}
    \caption{Foursquare New York City Daily Trajectory Number and User Trajectory Length (count by day).\vspace{-0.1cm}}
    \label{fig:rel_nyc}
    \vspace{-0.1cm}
\end{figure}

\begin{figure}[t]
    \centering
    \includegraphics[width=1\linewidth,trim=0cm 0cm 0.0cm 0.0cm,clip]{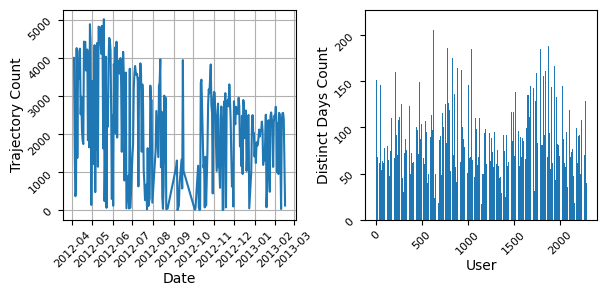}
    \vspace{-0.1cm}
    \captionsetup{width=1.\linewidth}
    \caption{Foursquare Tokyo Daily Trajectory Number and User Trajectory Length (count by day).\vspace{-0.1cm}}
    \label{fig:rel_tky}
    \vspace{-0.1cm}
\end{figure}

 The distribution of daily trajectory counts of this dataset 
is shown in Figure~\ref{fig:rel_nyc} for New York City and in Figure~\ref{fig:rel_tky} for Tokyo.
In this dataset, we observe a denser distribution of PoIs, along with a significantly larger user base—comprising 
1,083 users in New York City and 2,293 in Tokyo. The average daily check-in count is 902 for New York City and 2,215 for Tokyo, respectively. Some summary data is shown in Figures \ref{fig:rel_nyc} and \ref{fig:rel_tky}. Compared to previous datasets, this collection offers a richer and more diverse representation of user mobility patterns, making it especially valuable for long-term, fine-grained trajectory prediction tasks.

\subsection{Human Trajectory Prediction}\label{sec:related_paper}
Human trajectory prediction has garnered considerable attention in recent years. In this section, we introduce several representative methods that utilize machine learning methods that are closely aligned with the focus of our study. 

Yang et al.~\cite{yang2018spatio} proposed the Recurrent-Censored Regression (RCR) model, which mitigates the scarcity of check-in data by enriching it with inferred potential visitors. These potential visitors are identified using matrix factorization techniques based on historical behavior and user features. Similar work was done by Lian et al.~\cite{lian2013collaborative} and Ye et al.~\cite{ye2013s} using mixed hidden Markov models and Liu et al.~\cite{liu2016predicting} with a Recurrent Neural Networks (RNN) model. 

Hang et al.~\cite{10.1145/3219819.3219902} analyzed Wi-Fi access logs from Purdue University as a dense check-in dataset to study student mobility and proposed a heterogeneous graph-based embedding model, Embedding for Dense Heterogeneous Graphs (EDHG). EDHG constructs a weighted heterogeneous graph that captures correlations among users, PoIs, time slots. Prediction is dependent on node similarity. 

Xu et al.~\cite{10003126} and He et al.~\cite{he2016inferring} proposed a temporal-context-aware framework for inferring individual human mobility from sparse check-in data by modeling time-sensitive individual-location interactions. The framework comprises several components: time-sensitive interaction modeling, PoI attribute modeling, geo-influence modeling and social influence to capture user's temporal mobility preferences, PoI semantics information and spatial correlations between PoIs. 
Both studies employ negative sampling strategies to ensure that model training remains effective and computationally efficient.

Ke et al.~\cite{ke2022group} proposed a group-based Multi-Features Move (GMFMove) method for human mobility prediction, which clusters users with similar movement patterns into overlapping groups—allowing each user to belong to multiple groups simultaneously. Their MFMove model comprises two main components: a long-term preference learning module and a short-term preference learning module. The short-term module employs three parallel LSTM networks to model user transition sequences at different abstraction levels: location, category, and grid level.

\vspace{-0.2cm}
\subsection{Batch Size Effects on Training Dynamics and Generalization} \label{sec:related_batch}
While large-batch training is often favored in today's deep learning applications for its computational efficiency and stable gradient estimates, it can be suboptimal in scenarios characterized by limited, sparse, and highly independent human trajectory datasets. Recent work has revisited the role of batch size in deep learning optimization and generalization, with mounting evidence supporting the benefits of small-batch training.
In this section we will discuss some research which demonstrates how adjusting batch size affects training performance.

Masters and Luschi~\cite{masters2018revisitingsmallbatchtraining} systematically investigated mini-batch stochastic gradient descent (SGD) optimization through experiments on CIFAR-10, CIFAR-100, and ImageNet using AlexNet and ResNet. They demonstrated that increasing mini-batch size reduces the range of learning rates for stable convergence. Mini-batches like size $m = 2$ to $m = 32$ consistently yielded the best test performance due to more frequent and up-to-date gradient updates.

Complementary findings by Keskar et al.~\cite{keskar2017largebatchtrainingdeeplearning} further highlight the advantages of small-batch training. 
Their work emphasizes that small batch sizes introduce beneficial stochastic noise into the optimization process, which allows the model to escape narrow, sharp minima in the loss surface—regions typically associated with poor generalization. 
This strategy enables the discovery of global minima, which contribute to improved generalization performance. 
Additionally, small-batch training enables more frequent parameter updates, resulting in more up-to-date gradient information and more stable convergence during training. \joon{Do we have more recent references regarding batch size effects? Two papers cited here were published in 2017 and 2018 (arXiv).}
\yl{have some, but not well cited  }

\subsection{Challenges of Non-IID Data}\label{sec:related_iid}
This non-IID structure introduces significant challenges for machine learning models. 
In real-world settings, data collected from IoT devices or shared sources often exhibits strong internal correlations, violating the IID assumption. Samples from the same user, device, or environment are not independent or identically distributed, making non-IID data the norm rather than the exception~\cite{10192897,dundar2007learning}.
In this subsection, we examine the limitations and potential drawbacks of training models on datasets with non-IID characteristics.

Dundar et al.~\cite{dundar2007learning} emphasized that when training samples exhibit correlations in either features or labels, models trained under the IID assumption are prone to learning distorted or biased relationships. In contrast, adopting a random effects modeling approach can lead to improved performance by explicitly modeling the variability across data subgroups and mitigating the effects of non-IID distributions.

Zhang et al.~\cite{9496155} also emphasized that standard stochastic gradient descent (SGD) algorithms become less effective in the presence of distributional inconsistency, as non-IID data across clients leads to biased global model updates. In particular, the aggregation of local models—each trained on skewed client-specific data—can cause the global model to deviate significantly from an optimal solution. This divergence results in a substantial decline in overall model performance. The authors further noted that local models often evolve in disparate directions due to data heterogeneity, ultimately diverging both from each other and from the intended global model.

In the human mobility domain, Zhu et al.~\cite{9359187} provided a direct trial on the Geolife dataset, emphasizing that users’ travel patterns are inherently heterogeneous, resulting in non-IID data distributions across users. To address this challenge, they proposed a secure public data sharing strategy. The core idea is to allow clients to augment their local datasets with samples drawn from a globally available public dataset, thereby making their local data distributions more balanced and representative. This augmentation approximates an IID-like condition, significantly improving both model performance and convergence when training on non-IID data—bringing results closer to those achievable under ideal IID conditions.

Recent research has shown a growing trend of grouping data by various attributes—such as user social relation, location to help the model better capture the spatio-temporal relation in the data. 
However, such groupings must be conducted with caution, as improper handling can introduce violations of the assumption of independent and identically distributed.

%% file: content/preliminary.tex
\begin{figure*}[t]
    \centering
    \includegraphics[width=1\linewidth,trim=0cm 9.2cm 2.5cm 0cm,clip]{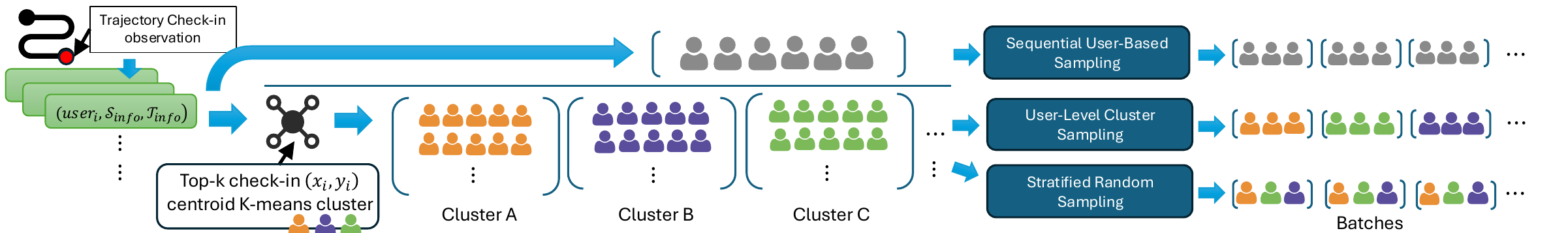}
    \vspace{-0.6cm}
    \captionsetup{width=1.\linewidth}
    \caption{User Personalized Representation and Sampling Module.\vspace{-0.45cm}}
    \label{fig:sampling}
    \vspace{-0.0cm}
\end{figure*}

In this section, we present the key concepts and definitions essential to our study on trajectory prediction used in the following sections.

\vspace{-0.2cm}
\subsection{Human Trajectory Representation}
A semantic trajectory is represented as a sequence of staypoints (or ``check-ins'') \( T = \{p_1 \rightarrow p_2 \rightarrow \dots \rightarrow p_n\} \) where $n$ is the number of staypoint observations in the trajectory.
Each stay point observation \( p_i \) in this sequence is a tuple containing three major parts \( (userID, \mathcal{S}_{info}, \mathcal{T}_{info} ) \),
$userID$ is a unique identifier of an individual user, $\mathcal{S}_{info}$ is the spatial information such as location $(x_i, y_i)$ in WGS84 format, travel distance $ds$ from previous check-in, and semantic information about the PoI category $cat$ \rk{v stands for what?, n is already related to POIs, if vn refers to a single POI, then it's confusing}. \yl{fixed}
$\mathcal{T}_{info}$ contains the necessary temporal information such as: day $d$, day of week $dow$, check-in time $t$, time segments $t_{seg}$, duration $du$.  The collection of all users' trajectories can be denoted as \( \mathcal{T} = \{T_1, T_2, \dots, T_m\} \), where \( m \) is the number of trajectories or users.

Depending on the dataset, not all attributes are directly available; however, some can be derived or generated from existing data. For example, FourSquare data provides information about the type of places (restaurant, school, etc.), whereas GeoLife data only provides raw geo-coordinates which can be used to detect staypoints but without any semantic information.

%
Table ~\ref{table:datasample} provides an overview of the spatial-temporal check-in record for three different FourSquare users. The first line of this example shows that this user stayed at a Cosmetics Shop with the given coordinates at 18:17 (UTC) April 3rd 2012.

%


\vspace{-0.2cm}
\subsection{Human Mobility Trajectory Prediction}
We consider the problem of human mobility trajectory prediction, where the objective is to forecast a user’s future movement behavior based on their historical mobility patterns. Specifically, we assume access to: (1) a training period, during which user trajectories are observed under typical conditions, and (2) a test period, during which the goal is to predict future trajectories based on patterns learned during training.

\begin{definition}[Train and Test Period]
Let \( \mathcal{T} = \{T_1, T_2, \dots, T_m\} \) be a set of trajectories of $m$ users.  We define a temporal cutoff point $t_{split}$, such that any check-in occurring before $t_{split}$ is used for training, while those occurring after are used for prediction. 
$$
\mathcal{T}_i^{Train}=\{(\mathcal{S}_{info},\mathcal{T}_{info} )[time]\in T_i| tc_i<=t_{split}\},
$$
and a testing trajectory:
$$
\mathcal{T}_i^{Test}=\{(\mathcal{S}_{info},\mathcal{T}_{info} )[time]\in T_i| tc_i>t_{split}\}.
$$
This allows us to define a train dataset as 
$$
\mathcal{T}^{Train}=[T^{Train}_1,...,T^{Train}_m],
$$
and a test dataset as
$$
\mathcal{T}^{Test}=[T^{Test}_1,...,T^{Test}_m].
$$
\end{definition}
Let $\mathcal{U} = \{u_1, u_2, \dots, u_m\}$ be the set of users, each associated with a training trajectory $\mathcal{T}_u^{Train}$. The task of trajectory prediction is to learn a model $$f_{\text{pred}}: \mathcal{T}_u^{Train} (\mathcal{S}_{info},\mathcal{T}_{info} )(x_i, y_i) \mapsto \hat{\mathcal{T}}_u^{Test}(\mathcal{S}_{info},\mathcal{T}_{info} )(x_i, y_i)$$ that estimates the sequence of future check-ins coordinates $(x,y)$ for each user over a specified prediction horizon.

This problem of human mobility trajectory prediction raises several important research questions in the training of trajectory prediction models.

\begin{enumerate}[noitemsep,
        nosep,
        leftmargin=10pt,
        labelsep=2pt,
        itemindent=0pt]
    \item \textbf{Question on Model Generalization:} Prior research has demonstrated that small-batch (small-batch stochastic gradient optimization) training often leads to improved generalization and training stability. However, relatively few studies have explored this effect in the context of sparse and individualized trajectory data, where each user’s mobility pattern is highly unique and irregular. This raises a critical question: How does batch size influence prediction accuracy when training models on such personalized spatio-temporal trajectories? 
    \item \textbf{Question on Semantic Information Selection:} Many recent models have incorporated contextual understanding of locations—such as the semantic categories of PoI as well as human life patterns to enhance predictive performance. While these features intuitively contribute to model accuracy, the relative importance of each remains underexplored. This raises a key question: Which type of semantic information contributes more significantly to trajectory prediction performance? 
    \item \textbf{Question on Dataset Sampling:} Human trajectories often exhibit interaction effects and spatial dependencies due to geographic co-location and shared behavioral patterns. As a result, the data may introduce frequently non-independent and identically distributed (non-IID), which poses challenges for model training and generalization. This raises an important question: How does the choice of sampling strategy—such as user-level cluster sampling, stratified sampling, or hierarchical clustering—affect model performance in the presence of non-IID data? 
\end{enumerate}
\andreas{This entire section reads very nicely. The definitions are sound. The example data is great. And challenges are well described. Very nice!}

%% file: content/methodology.tex
\begin{figure*}[t]
    \centering
    \includegraphics[width=.99\linewidth,trim=0.7cm 5.3cm 7.8cm 0.4cm,clip]{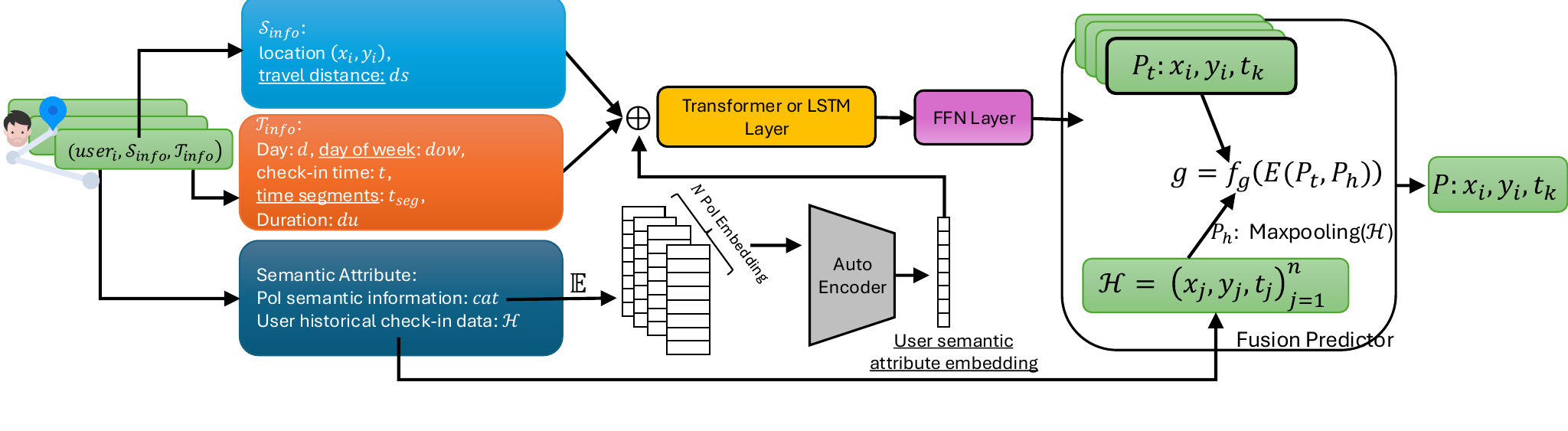}
    \vspace{-0.0cm}
    \captionsetup{width=1.\linewidth}
    \caption{The Illustration of Our Human mobility Prediction Framework. (Left) Spatio-temporal extraction from user check-ins; (Middle) LSTM/Transformer-based sequence model; (Right) Prediction via fusion with historical check-in \andreas{What's ``Trajectory Generator? We're using real-data arent we? Isn't the third box (two after Raw check-in data) still raw data?''}.\vspace{-0.45cm}}\yl{it used to generate extra feature dow/ user  life pattern }
    \label{fig:framework}
    \vspace{-0.0cm}
\end{figure*}

To address the aforementioned research questions, we introduce our modular Human Mobility Trajectory Prediction framework. The proposed model is designed with a modular architecture that enables flexibility, extensibility, and targeted analysis. 

\vspace{-0.3cm}
\subsection{Dataset Sampling}\vspace{-0.1cm}
To evaluate the IID characteristics of the human trajectory dataset, we employ a sampling module designed to assess how data distribution affects model performance.
In our test data, user identities are unknown, and it is unclear whether the underlying distribution is IID or non-IID, as shown in Figure \ref{fig:sampling}.
To address this, we introduce three sampling strategies: 
\begin{enumerate}[noitemsep,
        nosep,
        leftmargin=10pt,
        labelsep=2pt,
        itemindent=0pt]
\item Sequential User-Based Sampling (ordered by original dataset): In this baseline setting, data is loaded in a sequential manner based on user identifiers.
\item User-level Cluster Sampling: Users are grouped based on behavioral similarity.
In detail for each user, we extract their top-$k$ most frequently visited PoIs based on historical check-in records. If fewer than $k$ unique locations are available, we repeat the last seen PoI to maintain a consistent feature dimension. Next, we compute the centroid of the top-k PoIs in geographic space to represent the user’s dominant movement region:
%
First, we define a weight function \(w_{ij}\) for user $i$'s \( j^{\text{th}}\) most visited PoI, where the weight corresponds to the visit count at that location. 
The weighted centroid \(\mathbf{c}_i\) for user $\mathcal{U}_i$ is computed as follows:
\[\mathbf{c}i = \frac{\sum_{j=1}^{k} w_{ij} \, \mathbf{p}{ij}}{\sum_{j=1}^{k} w_{ij}}\]
where \(\mathbf{p}{ij} \in \mathbb{R}^2\) represents the latitude-longitude coordinates of user $\mathcal{U}_i$’s \(j^{\text{th}}\) most frequently visited location.
We then apply the K-means clustering algorithm on the resulting user centroids to group users into behaviorally similar clusters. The optimal number of clusters $K$ is determined through a brute-force search, using the silhouette coefficient as the selection criterion to evaluate clustering quality.
Users are ordered based on their assigned clusters and fed into the model accordingly. This ordering strategy is designed to enhance the effects of data shift and emphasize the non-IID characteristics of the dataset during training.

\item Stratified Sampling:
This approach is particularly effective for non-IID concern. In this strategy, users are first grouped into clusters based on their behavioral similarity, as determined by the user-level clustering process. Stratified sampling then constructs each training batch by selecting users proportionally from multiple clusters, rather than drawing samples arbitrarily or from a single cluster.
The primary goal of stratified sampling is to preserve inter-cluster variability during training, thereby provide a less skewed or context-dependent dataset. 
\end{enumerate}


\subsection{Comparative Studies}
In our comparative studies, we evaluate the impact of various contextual and temporal features extracted by the Trajectory Generation Module. This module processes raw tabular trajectory data to generate semantically rich representations. Specifically, we incorporate the following features:
\begin{enumerate}[noitemsep,
        nosep,
        leftmargin=10pt,
        labelsep=2pt,
        itemindent=0pt]
\item Day of the week, to capture periodic mobility patterns.
\item Travel distance from the previous PoI, which reflects travel mode.
\item Duration since the last check-in, indicating temporal gaps between two check-in.
\item Time segment classification, where each check-in is labeled as occurring during rush hour or non-rush hour.
\end{enumerate}
In addition to handcrafted features, we also incorporate semantic embeddings of users, generated using a pre-trained language model. These embeddings are derived from the textual descriptions of previously visited locations—such as venue category names—captured in the user's historical check-in data independently. 
This allows the downstream sequence prediction model to encode user preferences and behavioral patterns, while also to learn latent semantic similarities among users with comparable mobility behaviors.

\subsection{Trajectory Sequence Prediction}
The model incorporates two distinct trajectory sequence learning modules—Transformer~\cite{vaswani2023attentionneed} and LSTM~\cite{10.1162/neco.1997.9.8.1735}—each designed to capture different aspects of user mobility. The framework allows for selecting either architecture based on requirements.
\yl{cite 2: both of them are ts prediction using xx model related} \joon{at least, insert citations and add references for Transformer and LSTM. If there is no space, please consider adding equations to the appendix.}

The overall architecture is illustrated in Figure~\ref{fig:framework}. The feature name with an underline represents optional components that are evaluated through comparative experiments.


\subsection{Historical Fusion Predictor}

To effectively integrate real-time trajectory dynamics with long-term periodic user behaviors, we propose the Historical Fusion Predictor, inspired by previous research \cite{10.1145/3411812, ye2013s} on human check-in regularity patterns, which shows that individuals exhibit strong regularities in their mobility behaviors. \lance{Fixed base on Ruochen's comment} This component adaptively selects the location prediction between two candidates: one derived from the current trajectory and the other from historical behavioral patterns, with an emphasis on same-time past check-in behavior.

Specifically, the trajectory module produces a predicted location-time tuple:
\(P_T = (\hat{x}_T, \hat{y}_T, \hat{t}_T)\),
which represents the predicted spatial coordinates and timestamp.

From the user’s historical check-in sequence, defined as
\(\mathcal{H} = \{(x_j, y_j, t_j)\}_{j=1}^n\),
we extract a representative historical candidate \(P_H = (x_H, y_H, t_H)\) using a temporal alignment strategy and max-pooling over semantically relevant past check-ins.
Both candidates are encoded into latent embeddings:
\(e_T = \text{Emb}(P_T)\), \(\quad e_H = \text{Emb}(P_H)\).
To determine which candidate to trust, we construct a gating input vector:
\(X = [(e_T, e_H)]\),
and compute a scalar gate value \(f_g \in [0, 1]\) using a learned linear projection followed by a sigmoid activation:
\(f_g = \sigma(W^\top X + b)\),
where \(W\) and \(b\) are learnable parameters.

The final location prediction is selected based on the gate value:

\[P_{\text{final}} =
\begin{cases}
P_T & \text{if } f_g > 0.5, \\
P_H & \text{otherwise}.
\end{cases}\]

\newpage

%% file: content/results.tex

In this section, we present a comprehensive evaluation of our model utilizing two publicly available real-world datasets. The subsequent subsections provide detailed descriptions of each dataset, including their spatial and temporal characteristics, as well as the specific preprocessing steps applied to prepare the data for analysis. Furthermore, we outline the key findings in relation to the three previously posed research questions, supported by various model configurations and experimental setups.

%
%
%
%
\input{content/dataset}

%
%
%
\vspace{-0.0cm}
\subsection{Experimental Settings}

In this section, we detail the experimental settings used to validate the core concepts introduced in the preceding sections. We present the detailed configurations, sampling strategies employed, as well as the evaluation metrics used to assess the performance of our proposed model. 

\vspace{-0.1cm}
\subsubsection{Non-IID Structure and Dataset Shift\label{appendix:comarison}}
To evaluate the model’s performance under non-IID structure and dataset shift, we design three distinct testing strategies:
\begin{enumerate}[leftmargin=*]
    \item Sequential User-Based Sampling (ordered by original dataset): In this baseline trial, we assume that there is no inter-user variance or structural heterogeneity in the data.
    \item User-Level Cluster Sampling: We implement this to aggravate the inherent non-IID structure present in human mobility data.
    \item Stratified Random Sampling Across Clusters: We apply this to mitigate the potential effects of non-IID distributions.
\end{enumerate}

\subsubsection{Training Stability Under Varying Batch Sizes}
To investigate training stability under varying batch sizes, we evaluate the human mobility prediction model across five different batch size settings: 4, 16, 64, 256, 512.
\subsubsection{Comparative Experiment on Spatial Temporal info}
We conduct a comparative experiment to assess the contribution of key components in our model. Specifically, according to check-in trends shown in Figure \ref{fig:fs_hour}, we test the model with and without two temporal segmentation conditions: Rush hour (07:00–22:00) and Off-peak/night hours (23:00–06:00). In addition, we examine the effects of incorporating day-of-week indicators and travel distance.

\subsubsection{Comparative Experiment on User Semantic Attribute}
In some cases, the datasets may provide explicit user semantic information such as age, gender, or income as supplementary metadata. However, such attributes are rarely available and are often excluded due to privacy concerns or data anonymization.
To address this limitation, we propose an alternative strategy: deriving implicit user semantics from their PoIs. Our hypothesis is that the category of venues a user frequently visits encodes rich behavioral signals, capturing the interaction between the user.
Specifically, we use the venue category names (approximately 200 categories in our dataset) and embed them using a pre-trained language model, where each category is represented as a 512-dimensional vector. For each user, we extract embeddings for their top-10 most frequently visited PoI types and input them into an autoencoder for dimensionality reduction. The output is a fixed-length 512-dimensional latent vector—serving as the user's semantic representation.

\vspace{-0.2cm}
\subsubsection{Evaluation Metrics}
To rigorously evaluate the performance of our methodology and the contributions from the comparative experiment, we utilize GEO-BLEU~\cite{10.1145/3557915.3560951}: an adaptation of the traditional BLEU score from the natural language processing domain. In contrast to Dynamic Time Warping (DTW), GEO-BLEU does not require full-sequence alignment, making it more robust to minor deviations and better suited for capturing partial correctness in spatio-temporal trajectory predictions.
Furthermore, we employ the prediction accuracy as an additional evaluation metric in some experiments.

\begin{figure}[t]
    \centering
    \includegraphics[width=1\linewidth,trim=0cm 0cm 0cm 0cm,clip]{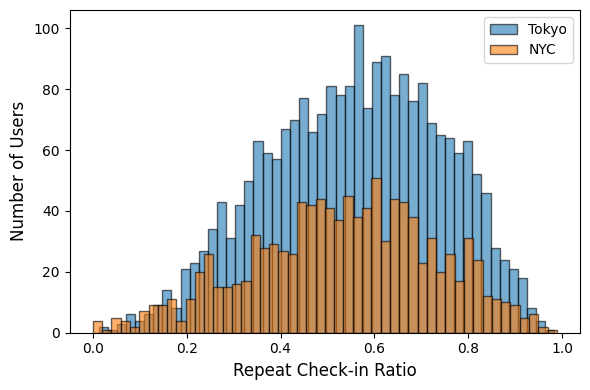}
    \vspace{-0.8cm}
    \captionsetup{width=1.\linewidth}    \caption{Average User Repeat Visit Ratio: NYC vs. Tokyo.\vspace{-0.45cm}}
    \label{fig:revisit}
    \vspace{-0.1cm}
\end{figure}

\vspace{-0.0cm}
\subsection{Prediction Results}
\subsubsection{Non-IID Structure and Dataset Shift Results}\label{sec:res_iid} For this task, experiments were performed using a Transformer model with a batch size of 4 and the Adam optimizer with a learning rate of 2e-5.
Table \ref{table:geo_bleu_sampling} demonstrates that stratified random sampling across clusters produces the superior performance in terms of GEO‐BLEU, achieving scores of 0.3374 in NYC and 0.2385 in Tokyo, which slightly outperforms sequential user‐based sampling with scores of 0.3341 (NYC) and 0.2367 (Tokyo). 
The similarity of the results of the sampling strategies suggests that sequentially iterating through the two datasets approximates a stratified sampling of local trajectory patterns, since the original dataset is not spatially ordered. Still, the stratified sampling approach ensures that each batch includes spatially diverse samples, whereas in some cases the sequential sampling may include spatially similar samples in the same batch by chance.

By contrast, user-level cluster sampling based solely on historical similarity of users yields remarkably low GEO‐BLEU scores. This aligns with our expectation, as this sampling method groups highly similar users into the same batch. It confines the model to a locally optimized solution and limits the generalizability of the model.
Across all sampling strategies, stratified random sampling proves particularly advantageous in non‐IID and on shifted datasets. Even when there is no clear evidence of heterogeneity in the test data, stratified random sampling across all clusters ensures the most comprehensive coverage of user behaviors.
\begin{table}[t]
    \centering
    \caption{Prediction Performance using Different Data Sampling Strategies. \vspace{-0.4cm}}
    \begin{adjustbox}{width=0.45\textwidth,center}
        \begin{tabular}{@{}lrrrr@{}}
            \toprule
            \multirow{2}{*}{\textbf{Sampling Strategies}} & \multicolumn{2}{c}{\textbf{GEO-BLEU}} & \multicolumn{2}{c}{\textbf{Accuracy}} \\
            \cmidrule(lr){2-3} \cmidrule(lr){4-5}
            & NYC & Tokyo & NYC & Tokyo \\
            \midrule
            Sequential User-Based Sampling & 0.3341 & 0.2367 &     10.12\% & \textbf{11.29}\% \\
            User-level Cluster Sampling & 0.2644 & 0.1979 &8.31\%  & 11.02\% \\
            Stratified Sampling Across Clusters & \textbf{0.3374} & \textbf{0.2385} & \textbf{10.17}\% & \textbf{11.29}\% \\
            \bottomrule
        \end{tabular}
    \end{adjustbox}
    \label{table:geo_bleu_sampling}
\end{table}


\subsubsection{Training Stability Under Varying Batch Sizes} \label{sec:res_batch}
To assess the stability and performance of our model under different batch size configurations, we conduct experiments using the transformer model, Adam optimizer with a learning rate of 2e-5, while varying the batch size across five settings: 4, 16, 64, 256, and 512. For this experiment, the model uses the Sequential User-Based Sampling as default.  

\begin{figure}[t]
    \centering
    \includegraphics[width=1\linewidth,trim=0cm 0cm 0cm 0cm,clip]{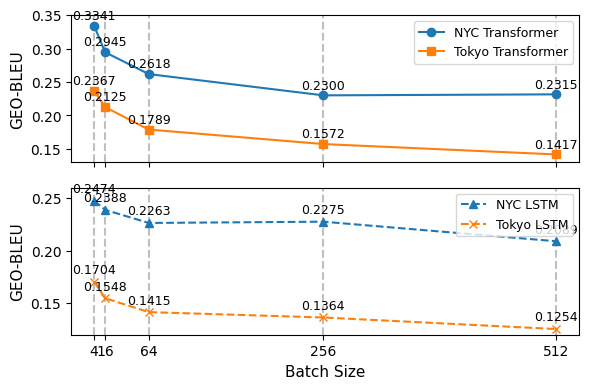}
    \vspace{-0.4cm}
    \captionsetup{width=1.\linewidth}
    \caption{Comparison of GEO-BLEU Scores Across Batch Sizes.\vspace{-0.45cm}}
    \label{fig:batch}
    \vspace{-0.1cm}
\end{figure}

As illustrated in Figure~\ref{fig:batch}, we observe that in human mobility datasets, model training stability and predictive performance seem highly sensitive to batch size. 
For the NYC dataset, the model achieves higher GEO-BLEU as the batch size decreases, with the highest performance (0.3341) at a batch size of only 4. Similarly, in the Tokyo dataset, the highest performance (0.2367) is also observed at the smallest batch size of 4. Thus, for predicting human mobility, we see that the additional computational cost of using a smaller batch size does increase prediction quality. In general, smaller batch sizes tend to provide better generalization performance, particularly in datasets with high behavioral variability, however, they significantly increase training time. On the other hand, larger batch sizes, such as 512, can accelerate training but require substantial computational resources, with memory consumption exceeding 60 GB. Therefore, selecting an appropriate batch size involves a trade-off between computational efficiency and robustness, and should be guided by both dataset characteristics and hardware constraints.

Another finding from our sampling experiments (\ref{sec:res_iid}) is that the variation in model performance across different sampling strategies was relatively small: only 1.86\% for NYC and 0.20\% for Tokyo in accuracy between the best and worst sampling configurations. Similarly, in the batch size experiments (\ref{sec:res_batch}), the accuracy metric remained largely unchanged across all settings. These observations provide two key insights:\newpage
\begin{enumerate}
    \item GEO-BLEU is more sensitive than accuracy for evaluating subtle differences in model performance, particularly in sequence prediction tasks with partial correctness.
    \item  In smaller datasets, due to limited coverage, the model may converge to memorizing or correctly predicting only the most regular patterns or “the lower bound” \cite{10.1007/s10707-016-0279-5}. As a result, accuracy can appear stable even when overall sequence quality degrades, whereas GEO-BLEU more effectively captures variations in prediction quality across the entire trajectory.
\end{enumerate}

\subsubsection{Comparative Study on Spatio-Temporal Datasets.}  
Table~\ref{table:performance} provides a comparative experiment to evaluate the impact of several components: time segmentation, user semantic attributes, and the Fusion Predictor Layer incorporating historical check-in frequencies. Specifically, we compare two spatio-temporal architectures (Transformer and LSTM) under the following settings: (1) with and without extra spatio-temporal info, (2) with and without user semantic attributes (historical check-in data), and (3) with and without the Fusion Predictor Layer that integrates historical check-in frequency information. As a baseline, we feed the models with only basic check-in information (the non-underlined attributes in Figure~\ref{fig:framework}), including location, check-in time, and duration.

\begin{table*}[t]
\Small
    \centering
    \caption{Prediction Performance. For each dataset, the highest GEO-BLEU scores are highlighted in bold.\vspace{-0.4cm}}
    \begin{adjustbox}{width=1\textwidth,center}
        \begin{tabular}{@{}lcccccccc@{}}
            \toprule
            & \multicolumn{4}{c}{{NYC}} 
            & \multicolumn{4}{c}{{Tokyo}} \\
            
            \cmidrule(lr){2-5} \cmidrule(l){6-9}
            
            & {Baseline}
            & {\makecell{Baseline with \\ext Spatio-temporal info}}
            & {\makecell{Baseline with User \\ semantic attribute}}
            & {\makecell{Baseline with \\Fusion Predictor Layer}}
            & {Baseline}
            & {\makecell{Baseline with \\ext Spatio-temporal info}}
            & {\makecell{Baseline with User \\ semantic attribute}} 
            & {\makecell{Baseline with \\Fusion Predictor Layer}}\\
            \midrule
            Transformer-Batch 4 & 0.3341 & 0.3437  & 0.1644  & \textbf{0.3547} & 0.2367 & 0.2386  & 0.1493  & \textbf{0.2675}\\
            Transformer-Batch 16 & 0.2945 & 0.2947  & 0.1601  & 0.3282         & 0.2125 & 0.2257  & 0.1379  & 0.2609\\ 
            Transformer-Batch 64 & 0.2618 & 0.2636  & 0.1645  &  0.3006        & 0.1789 & 0.1785  & 0.1068  & 0.2447\\
            Transformer-Batch 256 & 0.2300 & 0.2318  & NA  & 0.2831            & 0.1572 & 0.1609  & NA      & 0.2330\\ 
            Transformer-Batch 512 & 0.2315 & 0.2315  & NA  & 0.2840            & 0.1417 & 0.1443  & NA      & 0.2255\\  \midrule
            LSTM-Batch 4  & 0.2474 & NA & NA & \textbf{0.2889}    & 0.1704 & NA & NA & \textbf{0.2369} \\
            LSTM-Batch 16  & 0.2388 & NA & NA & 0.2837          & 0.1548 & NA & NA & 0.1979 \\
            LSTM-Batch 64  & 0.2263 & NA & NA & 0.2780          & 0.1415 & NA & NA & 0.1861 \\
            LSTM-Batch 256 & 0.2285 & NA & NA & 0.2804         & 0.1364 & NA & NA & 0.1816 \\            
            LSTM-Batch 512 & 0.2275 & NA & NA & 0.2809         & 0.1254 & NA & NA & 0.1783 \\
            \midrule
            HV\cite{kong2024human}          & 0.2089 & NA  & NA  & NA            & 0.1979 & NA  & NA  & NA \\
            \bottomrule
        \end{tabular}
    \end{adjustbox}
    \label{table:performance}
\end{table*}

For the Transformer-based model, the baseline GEO-BLEU score is 0.3341 for NYC and 0.2367 for Tokyo for the optimal batch size of 4. Incorporating extended temporal segmentation results in a modest performance improvement in both cities, indicating that distinguishing between rush and non-rush hours contributes to capturing periodic mobility patterns.
The addition of the Fusion Predictor
Layer, derived from users’ check-in histories and embedded into a latent representation, results in a larger performance boost in NYC, suggesting that personalized behavior significantly improves prediction accuracy. 
Although the standalone effect of semantic attributes is more muted in Tokyo, the integration with the Fusion Predictor Layer still raises performance to 0.2675, the highest across all configurations.
The LSTM model follows a similar trend, though with slightly lower overall performance, confirming that Transformer-based models better capture long-range dependencies in human mobility trajectories.
The Historical Visit (HV) model~\cite{kong2024human}, included for comparison, relies on a simple heuristic: it replicates the user’s most recent check-in sequence from a previous period to predict future trajectories. While this method can capture repetitive behavioral patterns, its performance is significantly lower than that of learning-based models, as it does not account for spatial context or dynamic temporal dependencies.


Interestingly, we observe that including semantic user attributes as an embedding of the user's PoI classes incurs a substantial loss in prediction accuracy. This is surprising because, intuitively, knowing the preferences of a user should help us predict the user's mobility. For example, knowing that a user likes to visit ``Coffee Shop'' PoIs should improve predictions by allowing predicting the user to go to areas with ``Coffee Shop'' or semantically similar PoIs. But we do not observe this in our experiments. We suspect that the large number of hundreds of different semantic PoI categories appears to introduce noise and uncertainty rather than a signal to improve the model’s predictive capabilities. Another reason for this lack of model improvement using semantic information is that the semantic labels of points of interest may be of low quality. 
Yet another reason could be that the exact location strictly contains more information than information about types of PoI. For example, if an user frequently visits a certain coffee shop, this user may not be more likely to visit other coffee shops, as they might prefer to get their favorite coffee rather than visiting other similar places. These theories aside, understanding why including semantic information about PoI categories and using the categories to describe semantic clusters of users remains an open problem.

%% file: content/dataset.tex
\begin{figure}[t]
    \centering
    \begin{subfigure}[t]{1\linewidth}
        \centering
        \includegraphics[width=\linewidth, trim=0cm 1.7cm 0cm 1cm, clip]{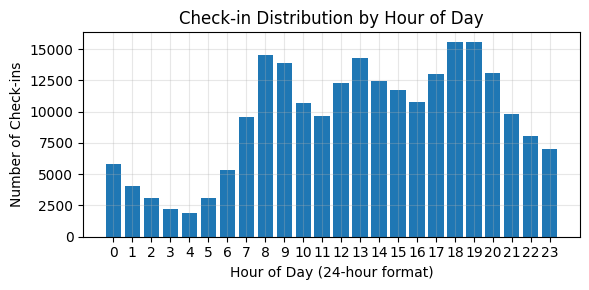}
    \end{subfigure}
    \begin{subfigure}[t]{1\linewidth}
        \centering
        \includegraphics[width=\linewidth, trim=0cm 0cm 0cm 1cm, clip]{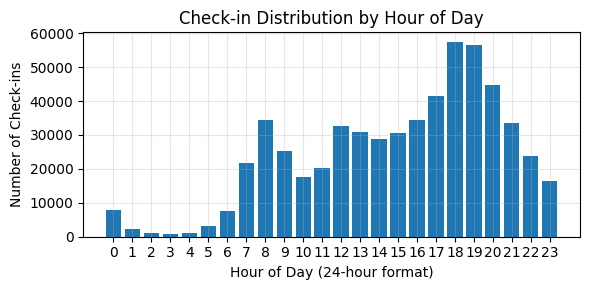}
    \end{subfigure}
    \vspace{-0.3cm}
    \captionsetup{width=1.\linewidth}
    \caption{Hourly Check-in Trends in New York (top) and Tokyo (bottom).}
    \label{fig:fs_hour}
    \vspace{-0.3cm}
\end{figure}

\vspace{-0.1cm}
\subsection{Experimental Datasets}
\label{sec:dataset}

Due to the inherent limitations of real-world trajectory datasets, we conducted an evaluation of dataset population coverage and trajectory density. Based on the assessment in Section~\ref{sec:related_db}, we selected the Foursquare check-in dataset~\cite{6844862} as the primary data source for our study.

\subsubsection{Data Analysis}

In this section, we investigate user behavioral patterns within the dataset to inform subsequent modeling steps. 
Specifically, we analyze users’ temporal visitation trends to uncover overarching daily routines and seasonal lifestyle patterns. 
For instance, we examine the most frequently visited categories of check-in locations, along with temporal patterns distributed across various hours of the day and days of the week. 

Our analysis of check-in frequencies across different times of day reveals a pronounced day–night cycle in both cities, characterized by significantly reduced activity during nighttime hours. As shown in Figure \ref{fig:fs_hour}, both cities encounter fewer check-ins occurring during nighttime hours. This consistent pattern indicates a shared rhythm of urban life structured around daylight-driven activity.

\begin{figure}[t]
    \centering
    \begin{subfigure}[t]{1\linewidth}
        \centering
        \includegraphics[width=\linewidth, trim=0cm 0.4cm 0cm 0cm, clip]{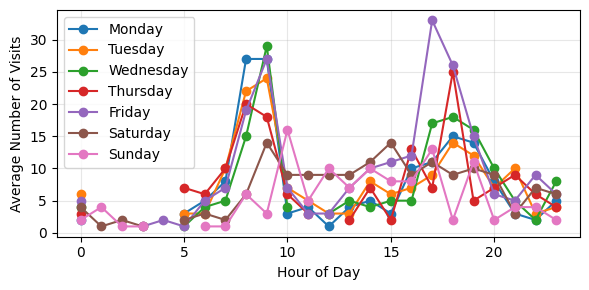}
    \end{subfigure}
    \vspace{-0.3cm}
    \begin{subfigure}[t]{1\linewidth}
        \centering
        \includegraphics[width=\linewidth, trim=0cm 0cm 0cm 0.3cm, clip]{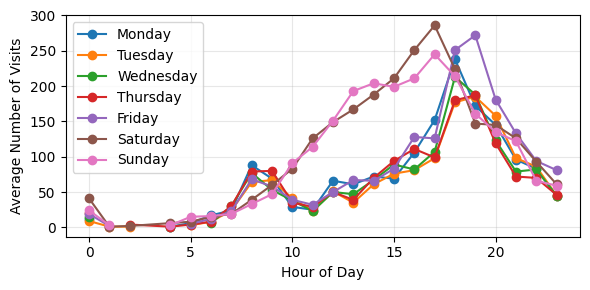}
    \end{subfigure}
    \vspace{-0.3cm}
    \captionsetup{width=1.\linewidth}
    \caption{Weekly Check-in Trends in New York (top) and Tokyo (bottom).\rk{This figure is never mentioned in the main text}}\yl{fixed}
    \label{fig:fs_most_busy_hour}
    \vspace{-0.3cm}
\end{figure}

To further explore temporal dynamics, we evaluate the weekly check-in distributions within the most active spatial cell in each city. 
Notably, in both cases, the busiest spatial cell corresponds to a major train station. As shown in Figure~\ref{fig:fs_most_busy_hour};
While both cities exhibit distinct dual-peak patterns—corresponding to morning and evening rush hours—New York experiences a more evenly distributed volume of check-ins throughout the entire day on weekends. 
The top 10 most frequently visited check-in venue categories in New York City and Tokyo reveal both similarities and distinctions in urban behavior, as shown in Table \ref{table:venue_categories}. While certain categories, such as “Train Station,” “Subway,” and “Coffee Shop”, appear prominently in both cities—reflecting common patterns of commuting and social activities—others differ, illustrating localized lifestyle preferences. For example, “Home (private)” and “Gym / Fitness Center” are more prevalent in NYC, whereas “Ramen / Noodle House” and “Convenience Store” feature more prominently in Tokyo.

Additionally, prior research has highlighted various aspects of spatiotemporal behavior modeling relevant to our study.
For example, Sadri et al. \cite{10.1145/3287064} extrapolates future locations by leveraging the corresponding time segments from historically similar days.
Another paper \cite{10.1145/3411812} explores various features—such as historical POI visitation frequency, contextual information and user-specific historical patterns. 
In light of these findings, we further investigate revisitation behavior in those two cities. Preliminary analysis reveals a strong recurrence of user check-ins to specific locations. To quantify this, we define a Repeat Check-in Ratio $RCR$ as:

$$\text{RCR} = \frac{TC_i - UC_i}{TC_i},$$
where $TC_i$ is the \underline{T}otal \underline{C}heck-in for user $\mathcal{U}_i$ and $UC_i$ is the \underline{U}nique \underline{C}heck-in for user $\mathcal{U}_i$.
This metric captures the degree of repetitiveness in user mobility. 
Figure \ref{fig:revisit} plots the distribution of this ratio, indicating that many users exhibit high revisitation tendencies.

\begin{table}[t]
    \centering
    \caption{Top 10 Most Frequent Check-in by Venue Categories.\vspace{-0.4cm}}
    \begin{adjustbox}{width=0.48\textwidth,center}
        \begin{tabular}{@{}lrlr@{}}
            \toprule
            \multicolumn{2}{c}{\textbf{NYC}} & \multicolumn{2}{c}{\textbf{Tokyo}} \\
            \cmidrule(lr){1-2} \cmidrule(lr){3-4}
            \textbf{Venue Name} & \textbf{Count} & \textbf{Venue Name} & \textbf{Count} \\
            \midrule
            Bar & 15,978 & Train Station & 200,428 \\
            Home (private) & 15,382 & Subway & 41,666 \\
            Office & 12,740 & Ramen / Noodle House & 17,303 \\
            Subway & 9,348 & Convenience Store & 16,833 \\
            Gym / Fitness Center & 9,171 & Japanese Restaurant & 15,680 \\
            Coffee Shop & 7,510 & Bar & 14,940 \\
            Food \& Drink Shop & 6,596 & Food \& Drink Shop & 14,023 \\
            Train Station & 6,408 & Electronics Store & 10,897 \\
            Park & 4,804 & Mall & 10,839 \\
            Neighborhood & 4,604 & Coffee Shop & 8,959 \\
            \bottomrule
        \end{tabular}
    \end{adjustbox}
    \label{table:venue_categories}
\end{table}

\subsubsection{Data Processing}
To ensure compatibility between the dataset and our model, and to mitigate noise that may arise from predicting precise latitude and longitude coordinates, we discretize the spatial domain by mapping each check-in location to a uniform grid. Specifically, we divide the geographic area into a 200 × 200 grid of spatial cells. For Tokyo, each cell measures approximately 199.7 meters in the east–west direction and 198.7 meters in the north–south direction. For New York City, each cell spans approximately 249.1 meters east–west and 243.5 meters north–south.

We partition the data temporally for training and testing. The training set comprises data from day 0 to day 249, while the test set covers day 250 to day 299. After applying the spatial discretization and temporal sampling, we obtain a total of 489,338 check-in records for Tokyo and 191,238 records for New York City.


%% file: content/conclusion.tex
In this study, we conducted several spatio-temporal trajectory prediction experiments with our framework that integrates temporal segmentation, user semantic attributes, and a fusion layer informed by historical check-in frequencies. Through comprehensive experiments on two real-world human mobility datasets, we demonstrated the effectiveness of each component via comparative experiments and performance evaluations under various data sampling strategies. Our results highlight the importance of modeling non-IID structures in mobility data and show that incorporating personalized behavioral features and temporal context substantially improves predictive accuracy.

Notably, we found that while the Fusion Predictor Layer consistently enhanced model performance, the inclusion of user semantic attributes based on historical check-in counts and PoI categories introduced uncertainty in some cases, potentially due to noisy or weakly informative embeddings. Additionally, our analysis on training stability under varying batch sizes revealed a clear trade-off between generalization and computational efficiency, reinforcing that batch size tuning must be dataset- and resource-aware.

In terms of data representation, our attempts to cluster PoIs using pre-trained semantic embeddings and K-means clustering yielded weak spatial separation, limiting their utility for downstream tasks. This suggests that semantic clustering alone may be insufficient without incorporating stronger spatial constraints or context-aware learning mechanisms.

\vspace{-0.0cm}

\begin{acks}
Supported by the Intelligence Advanced Research Projects Activity (IARPA) via Department of Interior/ Interior Business Center (DOI/IBC) contract number 140D0423C0025. The U.S. Government is authorized to reproduce and distribute reprints for Governmental purposes notwithstanding any copyright annotation thereon. Disclaimer: The views and conclusions contained herein are those of the authors and should not be interpreted as necessarily representing the official policies or endorsements, either expressed or implied, of IARPA, DOI/IBC, or the U.S. Government.
\end{acks}

\vspace{-0.0cm}